\pdfoutput=1

\documentclass[11pt]{article}

\usepackage[]{ACL2023}

\usepackage{times}
\usepackage{latexsym}

\usepackage[T1]{fontenc}

\usepackage[utf8]{inputenc}

\usepackage{microtype}

\usepackage{inconsolata}

\usepackage{algorithm}
\usepackage{algorithmic}
\usepackage{amsmath}
\DeclareMathOperator*{\argmax}{argmax}
\usepackage{times,latexsym}
\usepackage{url}
\usepackage[T1]{fontenc}
\usepackage{graphicx}
\newcommand{\bm}{\mathbf}
\usepackage{amssymb}
\usepackage{bbm}
\usepackage{booktabs}
\usepackage{xcolor}
\newcommand{\jr}{\textcolor{black}}

\newcommand{\gab}{\textcolor{black}}

\title{Event Knowledge Incorporation with Posterior Regularization for Event-Centric Question Answering}

\author{Junru Lu$^1$, Gabriele Pergola$^1$, Lin Gui$^2$ and Yulan He$^{1,2,3}$ \\
  $^1$Department of Computer Science, University of Warwick, UK\\
  $^2$Department of Informatics, King's College London, UK\\
  $^3$The Alan Turing Institute, UK\\
    \texttt{\{Junru.Lu, Gabriele.Pergola\}@warwick.ac.uk} \\
  \texttt{\{lin.1.gui, yulan.he\}@kcl.ac.uk}}

\begin{document}
\maketitle
\begin{abstract}
We propose a simple yet effective strategy to incorporate event knowledge extracted from event trigger annotations via posterior regularization to improve the event reasoning capability of mainstream question-answering (QA) models for event-centric QA. In particular, we define event-related knowledge constraints based on the event trigger annotations in the QA datasets, and subsequently use them to regularize the posterior answer output probabilities from the backbone pre-trained language models used in the QA setting. We explore two different posterior regularization strategies for extractive and generative QA separately. For extractive QA, the sentence-level event knowledge constraint is defined by assessing if a sentence contains an answer event or not, which is later used to modify the answer span extraction probability. For generative QA, the token-level event knowledge constraint is defined by comparing the generated token from the backbone language model with the answer event in order to introduce a reward or penalty term, which essentially adjusts the answer generative probability indirectly. We conduct experiments on two event-centric QA datasets, TORQUE and ESTER. The results show that our proposed approach can effectively inject event knowledge into existing pre-trained language models and achieves strong performance compared to existing QA models in answer evaluation.\footnote{Code and models can be found: \url{https://github.com/LuJunru/EventQAviaPR}.}
\end{abstract}

\section{Introduction}
\begin{figure}[h!]
  \centering
  \includegraphics[width=\linewidth]{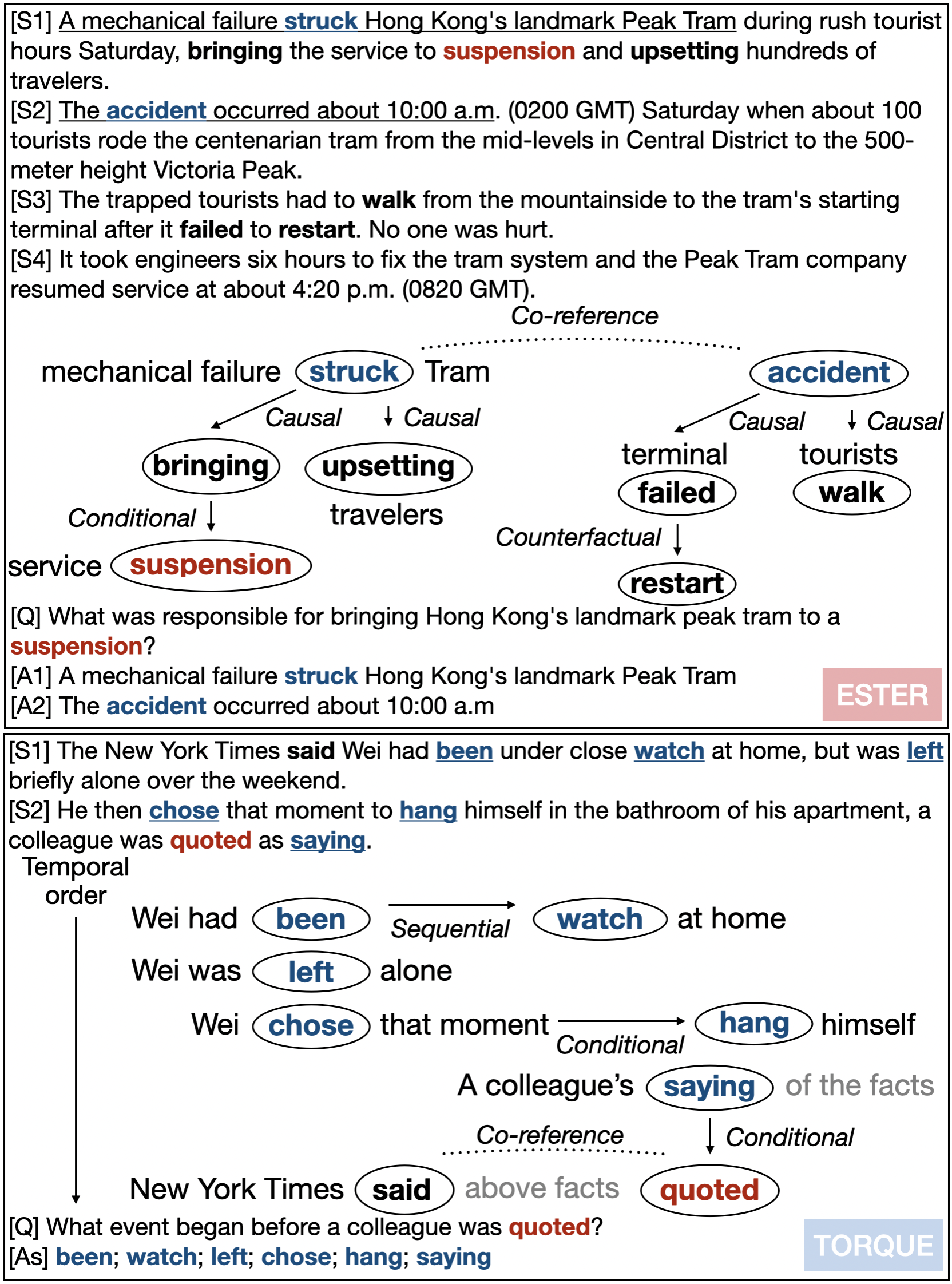}
  \caption{Event-centric QA examples from ESTER (top) and  TORQUE (bottom). All event triggers are in bold. The question event trigger and answer event triggers are further smeared with red and blue colors, respectively. In-text answers are highlighted with underlines. Event graphs are manually created for easy inspection of event-event relations.}
  \label{fig:sample}
\end{figure}

Question answering (QA) has been extensively explored on entity-centric corpora where QA pairs are centered on knowledge about entities occurred in text \cite{devlin2018bert,liu2019roberta,joshi2020spanbert}. More recently, attempts have been made to develop QA models requiring reasoning of event semantic relations such as temporal, causal, and event hierarchical relations \cite{souza2020event}. In general, an event can be defined as a key description word, often called an \emph{event trigger}, connected with a series of arguments \cite{zhang2020aser}. Events in text, which may or may not reside in the same sentence, could form various semantic relations. 
Two typical examples from existing event-centric QA datasets \cite{han2021ester,ning2020torque} are shown in Figure \ref{fig:sample}, in which a text paragraph is paired with a question and one or more answers. In these datasets, event triggers are annotated in the paragraphs, questions and answers. For easy inspection, we manually create an event graph for each example in Figure \ref{fig:sample}, where nodes denote events and edges show the semantic relation between two events. 
The first example from the ESTER dataset \cite{han2021ester} reports an event of `\emph{Peak Tram mechanical failure}' and its consequence.
The question asks about the cause which leads to the event of `\emph{suspension}'. To answer the question, a QA model needs to first understand the event semantic relation that the question cares about (i.e., `\emph{Causal}'), then identify the causal relation between the `mechanical failure \emph{struck} Peak Tram' event and the `\emph{suspension}' event, as well as the co-reference relation between `\emph{struck}' and `\emph{accident}', and finally provide correct answers.
In the second example, the question concerns events holding the temporal relation (i.e., `\emph{Before}') with the event `a colleague was \emph{quoted}'. To answer that, a QA model needs to be able to capture the temporal event knowledge in the given paragraph.

The main challenge of event-centric QA is the need to perform reasoning of event semantic relations in the given context. This is more difficult compared to entity-centric QA which typically only relies on statistical correlations between the question entities and the entities occurred in text, both encoded by pre-trained language models (PLMs). 
Also, PLMs can be easily tuned to learn entity knowledge in a self-supervised manner \cite{devlin2018bert,joshi2020spanbert}, but it is far more challenging to encode complex event semantic knowledge in PLMs. 

To address the challenges, we propose a simple yet effective strategy to incorporate event semantic knowledge via posterior regularization for both extractive and generative event-centric QA. In specific, event-related knowledge constraints are first defined based on the event trigger annotations in the QA datasets, which are subsequently used to regularize the posterior answer output probabilities from the backbone PLMs used in the QA setting. For extractive QA, we define sentence relevance by assessing if it contains an event trigger and if any of its event triggers is an answer event. The sentence relevance can be considered as the sentence-level event knowledge constraint which can be used as a regularized score to adjust the probability of extracting an answer span from the text. Intuitively, if the sentence relevance score is high, then the probability of extracting an answer span from the corresponding sentence will be increased. As for generative QA, we define the token-level event knowledge constraints by comparing the generated token from a PLM with the answer event to incur a reward if there is a match and a penalty for irrelevant event tokens. Directly adjusting the answer generative probabilities using the token-level event knowledge constraints would lead to unstable results in our experiments. We instead define an additional loss term based on our introduced reward \& penalty term. The answer generative probabilities are regularized in an indirect manner. It is worth mentioning that although our model training relies on event trigger annotations to define event knowledge constraints, such annotations are not necessary during the inference. 

Our contributions can be summarized as follows: \textbf{(1)} We present a simple yet effective posterior regularization mechanism for event-centric QA. The event knowledge constraints are defined based on the event trigger annotations in the training set, which are subsequently used to adjust the answer probabilities from the backbone PLMs in the QA. \textbf{(2)} We explore two different posterior regularization strategies for extractive and generative QA separately. For extractive QA, the sentence-level event knowledge constraint is defined by assessing if a sentence contains an answer event or not
. For generative QA, the token-level event knowledge constraint is defined by comparing the generated token from the backbone PLM with the answer event in order to introduce a reward or penalty term
. 
\textbf{(3)} We conduct experiments on two event-centric QA datasets containing different QA forms with a range of event semantic relations. To the best of our knowledge, this work represents a first attempt to incorporate event knowledge via posterior regularization for event-centric QA. We outperform strong baselines under both the extractive and generative settings on the Exact Match (EM) scores.

\section{Related Work}

\paragraph{Event-Centric Question Answering} Event-centric Question Answering has recently attracted increased attention in the research community \cite{jin2020forecastqa,zhou2019going,shang2021open}. Some work focused on questions concerning event temporal relations. ForecastQA \cite{jin2020forecastqa} organized the event-centric questions and answers in a multi-choice framework, and provided explicit event timestamps for answer selection. 
The TORQUE dataset \cite{ning2020torque} 
contains the temporal event questions under the extractive setting, requiring an accurate understanding of subtle, and at times implicit language nuances of temporal keywords. 
\citet{shang2021open} developed a custom model, OTR-QA, for the aforementioned TORQUE dataset. The OTR-QA model reformulates the temporal event-centric QA as an open temporal relation extraction task, incorporating contrastive loss to encode small temporal differences. Instead of focusing on temporal event relations only, 
\citet{du2020event} and 
\citet{liu2020event} proposed employing QA for event trigger and argument extraction. \citet{han2021ester} developed an event-centric QA dataset, called ESTER, consisting of questions on five different event semantic relations. They also leveraged PTMs for QA under both the generative and extractive settings. \jr{\citet{lu2022event} proposed to transform the contextual embeddings of events into an event-centric space, and utilize contrastive learning to incorporate event knowledge into event-centric QA on the ESTER dataset. Their approach is only able to inject token-level event knowledge. In contrast, our method provides a more flexible way of fusing event knowledge with constraint functions tailored to the extractive and generative setting separately.}

\begin{figure*}[t]
  \centering
  \includegraphics[width=0.95\linewidth]{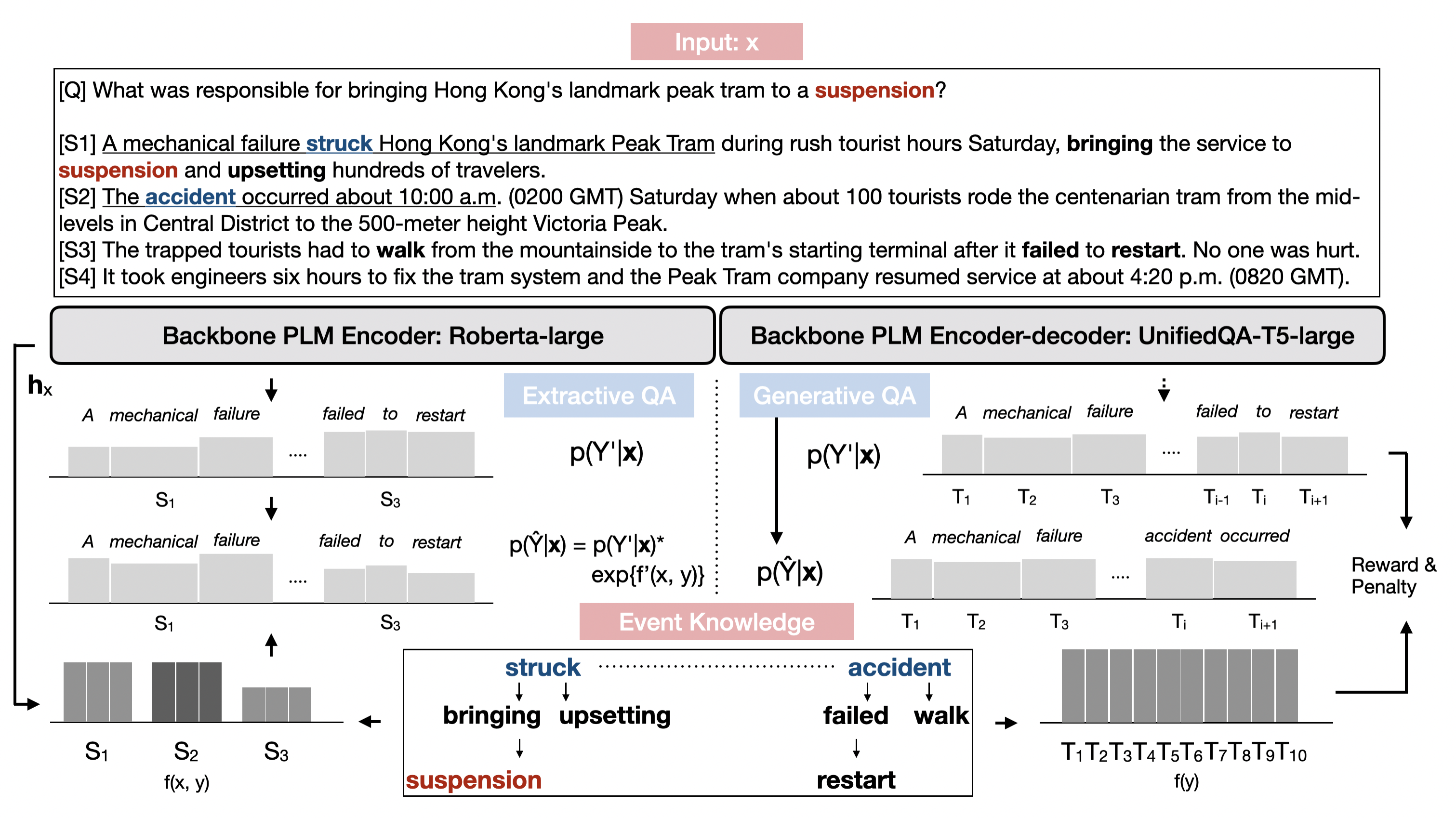}
  \caption{The overall architecture of our proposed mechanism. The input to the backbone PLM model is the question and the given supporting paragraph (upper box). In  \textbf{extractive QA (bottom left)}, we first use hidden vectors $\bm{h_x}$ to predict the sentence relevance probability $f'(x, y)$ with the true knowledge constraint $f(x, y)$ as the reference, which is then combined with the preliminary probability $p(Y'|\bm{x})$ to generate the regularized probability $p(\hat{Y}|\bm{x})$. In \textbf{generative QA (bottom right)}, the preliminary probability $p(Y'|\bm{x})$ and the knowledge constraint $f(y)$ are used to compute the reward\&penalty term, which rewards the generation of relevant triggers and penalises the generation of irrelevant event triggers. An additional loss term based on the reward\&penalty term will then indirectly guide the backbone PLM to generate the regularized probability $p(\hat{Y}|\bm{x})$.}
  \label{fig:model}
\end{figure*}

\paragraph{Posterior Regularization} Posterior regularization (PR) \cite{ganchev2010posterior} is a widely adopted framework to smoothly apply external knowledge constraints. 
\citet{ganchev-das-2013-cross} and 
\citet{zhang2018prior} fused PR with machine translation, while 
\citet{yang2014context} and 
\citet{zhao2016semi} explored the application of PR to sentiment classification. 
\citet{zhou2020robust} constructed entity, lexical, and predicate knowledge constraints to enhance QA models in identifying tiny language differences in adversarial perturbations on entity-centric QA datasets. To the best of our knowledge, this work represents the first attempt to design a posterior regularization mechanism built on event knowledge for event-centric QA.

\section{Methodology}
We first formulate the task of event-centric QA, and then introduce our proposed methodology.

\subsection{Task Formulation}
We formulate event-centric QA as a typical QA task with additional supervision signals of event triggers annotated in text passages, questions and answers. More formally, for a text passage $\bm{x}^{p}$ paired with an answerable event-centric question $\bm{x}^q$, a QA model is expected to produce one or more answers $\hat{Y}=\argmax_Y p(Y|\bm{x}^{q}, \bm{x}^{p})$, where $\hat{Y}=\{\bm{\hat{y}}_1, \cdots, \bm{\hat{y}}_A\}$, and $A$ denotes the number of answers. In addition, events are annotated in each text passage, their associated question, and ground truth answers, $E = \{\bm{e}_1^p,\cdots,\bm{e}_{C_p}^p,\bm{e}_1^q,\cdots,\bm{e}_{C_q}^q,\bm{e}_1^a,\cdots,\bm{e}_{C_a}^a\}$, where $C_p$, $C_q$ and $C_a$ denote the number of events in a passage, question and answers, respectively.

We build our QA models on the TORQUE \cite{ning2020torque}  and the ESTER \cite{han2021ester} datasets. In both datasets, event triggers are annotated and each question contains a single event trigger (i.e., $C_q=1$). In the subsequent sections, we use \emph{events} and \emph{event triggers} interchangeably. TORQUE focuses on the event temporal relation only and all its answers are just a set of event triggers holding the desirable temporal relation with their respective question events. ESTER covers a range of event semantic relations including \emph{Causal}, \emph{Conditional}, \emph{Counterfactual}, \emph{Subevent}, and \emph{Co-reference}. In ESTER, each question is additionally annotated with the event relation type $t$. Although answers provided in ESTER are just text spans from their respective text passages, the QA models can be developed under both the extractive and generative settings. 

\subsection{Injecting Event Knowledge via Posterior Regularization}
We propose an effective way to inject event knowledge via posterior regularization (PR) for both extractive and generative event-centric QA. \jr{Let $\theta$ denote the parameters of the basic QA model. Meanwhile, we define a set of PR constraints $f(\bm{x},\bm{y})$ built upon an input $\bm{x}$, comprising of a question $\bm{x}^q$ and a text passage $\bm{x}^p$, and a reference answer $\bm{y}$. The overall learning objective is defined by:
\begin{equation}\small
  J(\theta,\gamma) = L(\theta) - \sum_{n=1}^{N}\mbox{KL}(p(\hat{Y}|\bm{x};\gamma,\theta)||p(Y'|\bm{x};\theta))\label{eq:original-obj}
\end{equation} 
where $\gamma$ denotes the additional parameters to encode PR constraints, $p(\hat{Y}|\bm{x};\gamma,\theta)$ denotes the regularized probability, and $p(Y'|\bm{x};\theta)$ denotes the original probability without regularization. Particularly, $p(\hat{Y}|\bm{x};\gamma,\theta)$ is obtained with the function $G$ applied on the learned PR constraints $f'(\bm{x},\bm{y};\gamma)$ and $p(Y'|\bm{x};\theta)$:
\begin{equation}\small
  p(\hat{Y}|\bm{x};\gamma,\theta) = G\big(f'(\bm{x},\bm{y};\gamma), p(Y'|\bm{x};\theta)\big)
  \label{eq:G-function}
\end{equation}
To enable end-to-end training, 
following the mutual distillation algorithm proposed in \cite{hu2016deep} which converts the learning of PR into an optimization problem, we propose to transform the learning objective defined in Eq. (\ref{eq:original-obj}) to:
\begin{equation}\small
  L(\theta,\gamma) = L_{QA}(p(\hat{Y}|\bm{x};\gamma,\theta)) + L_{PR}(G;\gamma,\theta))
\end{equation}
}
The overall architecture is shown in Figure \ref{fig:model}. We take the PLM encoder, RoBERTa-large \cite{liu2019roberta}, and the PLM encoder-decoder, UnifiedQA-T5-large \cite{khashabi2020unifiedqa}, as the backbone for extractive and generative QA respectively. The input to our QA models is a question-paragraph pair shown in the upper box in Figure \ref{fig:model}. In the bottom part, event knowledge constraints are defined based on the event trigger annotations in the training set, which are subsequently used to regularize the conditional probability of the generated answer. In what follows, injecting event knowledge via posterior regularization will be discussed for both extractive and generative QA.

\subsubsection{Event-Centric Extractive QA}
In extractive QA, answers are either event trigger words in the text passages in TORQUE or text spans in ESTER. The problem can be framed as sequence labeling that given an input text passage, a QA model aims to produce a label sequence with a label assigned to each word token. Either the `B-I-O' tagging or the `I-O' tagging can be used, where `B' and `I' represent the beginning and the body of an answer span. In our experiments, we found that the `I-O' tagging works better. Therefore, the `I-O' labeling is used for both the TORQUE and ESTER datasets. We use $\bm{x} = \{$<s>$\bm{x}^{q}$</s></s>$\bm{x}^{p}\}$ to denote the input $\bm{x}$ to the RoBERTa-large encoder, in which `<s>' and `</s>' are special separator tokens. For the ESTER dataset, the event relation label $t$ is also given for each question. It can be used as a prompt, thus the input $\bm{x}$ for a training instance in ESTER is slightly modified as $\bm{x} = \{$<s>t:$\bm{x}^{q}$</s></s>$\bm{x}^{p}\}$. The target label in our setting is $Y=\{\bm{y}_1, \cdots, \bm{y}_{|\bm{x}^{p}|}\}$, where $\bm{y}_i$ is either `I' or `O'. 
Let $N_{x}$ be the total length of input sequence $\bm{x}$, $d$ be the dimension of the hidden representation produced by the RoBERTa-large, $\bm{h_x}\in\mathbb{R}^{N_{x}\times d}$, $Y'$ and $p(Y'|\bm{x})\in\mathbb{R}^{N_{x}}$ are the hidden representation of the input $\bm{x}$, and the probabilities of the preliminary answer label sequence $Y'$ before posterior regularization, respectively.

\noindent\underline{\textbf{Knowledge constraints}} 
We define the sentence-level event knowledge constraint $f(x_s,y)$:
\begin{equation}\small
  f(x_s, y) = \left\{
  \begin{array}{rcl}
      1 & & e^{a} \in x_s\\
      -1 & & \mbox{Otherwise}\\
  \end{array}
  \right.
\end{equation}
where $e^{a} \in x_s$ denotes that an answer event trigger word $e^{a}$ locates in sentence $x_s$. During training, we need to automatically infer a regularization score $f'(x_s, y)$ for each sentence $x_s$. It can be done by first deciding if an event trigger word $e_k$ in the sentence $x_s$ is an answer event and then taking the weighted aggregation of the classification results of all the event trigger words in the sentence. More concretely, for an event $e_k$ in the sentence $x_s$, we predict if it is an answer event by:
\begin{equation}\small
  g'(e_k) = \sigma(W_{g}^\intercal\bm{e}_{k} + b_{g}),
  \label{eq:etype}
\end{equation}
\noindent where $\sigma(\cdot)$ denotes the sigmoid function, $\bm{e}_{k}\in\mathbb{R}^d$ is the hidden representation of the event $e_k$ generated by RoBERTa-large, $W_{g}\in\mathbb{R}^{d}$ and $b_{g}$ are trainable weights and bias. 
The ground truth $g(e_{k})$ is defined as:
\begin{equation}\small
  g(e_{k}) = \left\{
  \begin{array}{rcl}
      1 & & e_{k} \in \{e^a_1, e^a_2, \cdots, e^a_{C_a}\}\\
      -1 & & \mbox{Otherwise}\\
  \end{array}
  \right.
\end{equation}
Essentially, $g(e_k)$ can also be considered as a constraint encoding the answer event information.

Assuming the sentence $x_s$ contains $K$ events, we then derive the regularized score $f'(x_s,y)$ by taking the weighted aggregation of the event classification results:
\begin{equation}\small
  f'(x_s, y) = \frac{1}{K}\sum_{k}\alpha_{\omega}^{k}g'(e_{k})
  \label{eqa:h}
\end{equation}
where $\alpha_{\omega}^{k}$ is the attention score of the $k$-th event in sentence $x_s$ with respect to question $x_q$, $\omega$ is the parameters of a multi-head attention (MHA) layer \cite{vaswani2017attention}:
\begin{equation}\small
  \mathbf{\alpha}_{\omega} = \mbox{MHA}(\bm{h}_{x_{s}},\bm{h}_{x_q},\omega) \label{eq:attention}
\end{equation}
\jr{To sum up, constraint $g(e_k)$ is defined at the token level to predict if a token is an answer event, while $f(x,y)$ is defined at the sentence level to predict if a sentence contains an answer event trigger word. The hierarchical constraints can encourage the major type of event triggers to contribute more to the prediction of the sentence regularization score $f(*)$. Such a formulation (e.g. a binary constraint indicating the presence or absence of certain knowledge) is commonly used in PR work \cite{zhou2020robust}.}

\noindent\underline{\textbf{Learning Objectives}} We can learn the sentence-level event knowledge constraints, $f'(x_s,y)$ and $g'(e_k)$, by minimizing the mean square error of the estimated scores and the ground truth scores:
\begin{equation}\small
  L_f=\frac{1}{S}\sum_{s}\big(f'(x_s,y) - f(x_s,y)\big)^2
\end{equation}
\begin{equation}\small
  L_g=\frac{1}{S}\sum_{s}\frac{1}{s_K}\sum_{k}\big(g'(e^k) - g(e^k)\big)^2
\end{equation}
where $S$ denote the total number of sentences, and $s_K$ denotes the total number of event triggers in sentence $s$. Once the sentence-level event knowledge constraint $f'(x_s, y)$ is learned, we can update the posterior answer probability by \jr{the following 
function $G$ mentioned in Eq. (\ref{eq:G-function}). The update of the $G$ function here is equivalent to optimizing the parameters $\alpha_{\omega}^k$ and the function $g'(e_k)$ in Eg. (\ref{eqa:h})}:
\begin{equation}\small
  p(\hat{Y}|\bm{x}) = G_{ext}(p(Y'|\bm{x},f'(\bm{x},Y))
\end{equation}
\begin{equation}\small
  G_{ext}(*) = p(Y'|\bm{x})\exp\{f'(\bm{x},Y)\}
\end{equation}
where $f'(\bm{x},Y)\in\mathbb{R}^{N_{x}}$ is the predicted regularization score for the whole input sequence. Note that for each sentence $x_s$, there is only a single value of $f'(x_s, y)$ computed. We populate this value for all the word tokens in the sentence $x_s$ and form a vector. Intuitively, if the score of $f'(x_s,y)$ is high, it indicates that the sentence $x_s$ more likely contains the answer event, as such, the conditional answer probability should be increased. In contrast, if the score of $f'(x_s,y)$ is low, then the sentence $x_s$ is more likely irrelevant and hence the conditional answer probability should be decreased. 

In the bottom left part of Figure \ref{fig:model}, the preliminary answer probability $p(Y'|\bm{x})$ mistakenly assigns a high score to Sentence 3 but a low score to Sentence 1. But the regularized probability $p(\hat{Y}|\bm{x})$ gives a correct prediction since the regularization constraint $f'(\bm{x},Y)$ guides the model to focus more on the first two sentences. The overall loss is:
\begin{equation}\small
  L_{ext} = L_{ext_{QA}} + \lambda_1(L_{f} + L_{g}) \label{eq:extqa}
\end{equation}
where $\lambda_1$ is the hyperparameter used to balance the training of the main task and the sub-tasks. $L_{ext_{QA}}$ is the loss of the main QA task:
\begin{equation}\small
  L_{ext_{QA}} = -\frac{1}{N_x}\sum_{i=1}^{N_x}w_{bl}y_{i}\log \hat{y}_{i}
  \label{eq:extl}
\end{equation}
where $w_{bl}$ is a balancing hyperparameter to boost the weights of positive labels \cite{han2021ester}.

\subsubsection{Event-Centric Generative QA} 

As there is no generative setting in the TORQUE-related work, we only explore generative QA on the ESTER dataset. We choose UnifiedQA-T5-large as the backbone encoder-decoder PLM, following the typical setup in the ESTER baseline that fine-tuning a T5-large \cite{raffel2019exploring} model in the universal generative style \cite{khashabi2020unifiedqa}. The input sequence $\bm{x}$ is denoted as $\bm{x}=\{t$:$\bm{x}^{q}\backslash n\bm{x}^{p}\}$ since T5 adopts text separators different from RoBERTa. The ground truth $Y$ is a concatenation of all answers separated by the `;' special token: $Y=\{\bm{y}_1; \cdots; \bm{y}_A\}$. Let $T$ be the total length of the output sequence $\hat{Y}$, $V$ be the size of the T5 vocabulary, $Y'$ and $p(Y'|\bm{x})\in\mathbb{R}^{{T}\times V}$ denotes the preliminary answer and its probability generated from the T5 decoder, respectively.

\noindent\underline{\textbf{Knowledge constraints}} Incurring penalties of certain tokens during generation via unlikelihood training is a popular strategy in controllable text generation \cite{welleck2019neural,devaraj2021paragraph,li2019don}. We extend this strategy by combining rewards of answer events and penalties of irrelevant events. Since the length of text generated by the decoder is unknown, we define the token-level event knowledge constraint $f(Y)$ for the generated text $Y$ as:
\begin{equation}\small
  f(y_i) = \left\{
  \begin{array}{rcl}
      \tau_1w_{\tau_1, \tau_2} & & y_i \in E^{p}, y_i \in E^{a}\\
      \tau_2w_{\tau_1, \tau_2} & & y_i \in E^{p}, y_i \notin E^{a}\\
      w_{\tau_1, \tau_2} & & y_i \notin E^{p}\\
  \end{array}
  \right.
  \label{eq:temp}
\end{equation}
\begin{equation}\small
    w_{\tau_1, \tau_2} = \frac{1}{|V|+(\tau_1 - 1)C_a+(\tau_2 - 1)(C_p - C_a)}
\end{equation}
where $E^p$ and $E^a$ denote events in the text passage and the answer, respectively, $E^p=\{e^p_1, e^p_2, \cdots, e^p_{C_p}\}$, $E^a=\{e^a_1, e^a_2, \cdots, e^a_{C_a}\}$, $\tau_1\in(0,1)$ and $\tau_2>1$ are hyperparameters to control the temperature of the softmax weights. The equation assures that the weighted sum of all tokens in the vocabulary equals 1. Conditional $\tau_1$ and $\tau_2$ ensure that the generation of an answer event will receive a reward while the generation of an irrelevant event will be penalized. $f(\bm{y})$ is created by applying $f(y_i)$ at all time steps during decoding. 

In the bottom right part of Figure \ref{fig:model}, the preliminary probability $p(Y'|\bm{x})$ mistakenly generates the second answer `\emph{failed to restart ···}'. This output does not align with the event knowledge constraint $f(y_i)$, and therefore receives a large penalty. The regularized probability $p(\hat{Y}|\bm{x})$ instead correctly generates the new answer `\emph{accident occurred ···}'.

\noindent\underline{\textbf{Learning objectives}} 
In the extractive QA setting, the event knowledge constraints are used to adjust the answer span extractive probability directly. However, in the generative QA setting, we found that directly modifying the posterior answer generative probability using the token-level knowledge constraints would lead to unstable results in our experiments. Therefore, we instead introduce a reward\&penalty term $r_{t}$ \jr{via a new $G$ function}, and define its associated loss term below:
\begin{equation}\small
  r_{t} = G_{gen}(f(v_j), p(y'_t=v_j|\bm{y}_{<t}, \bm{x})\big)
\end{equation}
\begin{equation}\small
  G_{gen}(*) = -f(v_j)\log\big(1-p(y'_t=v_j|\bm{y}_{<t}, \bm{x})\big)\label{eq:reward}
\end{equation}
\begin{equation}\small
  L_{RP} = \frac{1}{T}\sum_{t=1}^{T} r_{t}
\end{equation}
where $p(y'_t=v_j|\bm{y}_{<t},\bm{x})$ is the preliminary probability assigned to token $v_j$ at $t$-th position given the input sequence $\bm{x}$ and the output text sequence generated so far $\bm{y}_{<t}$. $f(v_j)$ denotes the token-level event knowledge constraint score for token $v_j$. $T$ denotes the total number of generated tokens in the model-output answer. \jr{We use the predefined constraint $f(y_i)$ in Eq. (\ref{eq:temp}) as the scaling weight with the preliminary probability $p(y'_t=v_j|\bm{y}_{<t},\bm{x})$ to create the unlikelihood term in Eq. (\ref{eq:reward}), which} ensures that the generation of an irrelevant event token receives a higher penalty score than any other tokens, while the generation of an answer event token receives a lower penalty score, and thus can be essentially considered as a reward.

The overall loss is defined as:
\begin{equation}\small
  L_{gen} = L_{gen_{QA}} + \lambda_2(L_{RP}) \label{eq:genqa}
\end{equation}
where $\lambda_2$ is a hyperparameter used to balance the loss term. $L_{gen_{QA}}$ is the loss of the main QA task:
\begin{equation}\small
  L_{gen_{QA}} = -\frac{1}{N_a + A - 1}\sum_{t=1}^{N_a + A - 1}y_{t}\log \hat{y}_{t}
\end{equation}
where $N_a$ is the total token length of $A$ ground truth answers separated by $A - 1$ `;' special tokens.

\section{Experiments}
\gab{We conduct a thorough experimental evaluation to assess the impact of the regularization strategies on the extractive and generative event-centric QA tasks. We first discuss the experimental setup, then the quantitative impact of the regularization strategy, and we finally conclude with a discussion of relevant case studies and an error analysis.}

\subsection{Experimental Settings}
\paragraph{Datasets} Two event-centric QA datasets are currently available in the literature: ESTER \cite{han2021ester} and TORQUE \cite{ning2020torque}. 

The ESTER dataset consists of over 1.9k TempEval3 (TE3) news snippets with annotated event triggers \cite{uzzaman2013semeval}, composed by extracting 3-4 consecutive sentences from news paragraphs with at least 7 event triggers. The dataset further includes 6k crowd-sourced answerable event-centric questions. For each question, ESTER provides question-answer event relations from a predefined label set: \emph{Causal}, \emph{Conditional}, \emph{Counterfactual}, \emph{Sub-event}, and \emph{Co-reference}, with the following proportion, respectively: 43.1\%, 21.3\%, 7.1\%, 15.6\% and 12.9\% proportions, respectively. With the exception of \emph{Sub-event} type questions having more than 3 answers on average, questions have generally 1-2 answers on average. 

The TORQUE dataset contains 30.4k temporal event-centric questions for over 3.2k news passages selected from the TE3 corpus. However, compared to ESTER, in TORQUE only two sentences were sampled to compose the final snippets. For each passage, TORQUE provides 3 hard-coded questions always enquiring about "past", "ongoing" and "future" events mentioned in text, and further additional user-generated questions querying the temporal relations between specific event triggers. 
Unlike ESTER whose answers are in the form of text spans, in TORQUE all the answers are lists of single event triggers occurring in text. Another difference regards the types of event questions, with ESTER covering semantic event questions and TORQUE focusing only on temporal event relations. 

The ESTER dataset has 4,547 training, 301 development and 1,170 test instances. The TORQUE dataset has 24,523 training, 1,483 development and 4,468 test instances. Due to the unavailability of ground-truth answers in the published test sets, the models are evaluated on the development sets.

\paragraph{Baselines} For the ESTER dataset, we report the original results of the fine-tuned RoBERTa-large model for the extractive QA, and the results obtained with the fine-tuned UnifiedQA-T5-large model for generative QA. \jr{We also include results from TranCLR \cite{lu2022event}}, the state-of-the-art model on ESTER. As for TORQUE, we only list the results available in the literature on the fine-tuned RoBERTa-large pipeline. 
Hyperparameters and training costs are reported in Appendix \ref{sec:hyperparams}.

\paragraph{Metrics} To generate comparable results, we use the same evaluation metrics adopted in the dataset papers. \citet{han2021ester} uses $F_{1}^{T}$, $HIT@1$ and $EM$ for both extractive and generative QA. $F_{1}^{T}$ considers overlaps of all unigrams between synthetic and the ground truth answers; $HIT@1$ detects if the top predicted answer contains the same event trigger as the leftmost golden answer, and Exact Match ($EM$) measures whether any synthetic answer matches exactly any reference answer of the ground truth set. \citet{ning2020torque} uses standard token-level macro $F_1$ and $EM$ metrics. In addition, they add an $EM$ consistency metric $C$ computing the percentage of contrast groups for which a model’s predictions have $F_1\geq$ 80\% for all the questions in a group. A higher $C$ score indicates a better distinction on contrast questions.

\begin{table}[t]
\resizebox{\columnwidth}{!}{%
  \begin{tabular}{l|ccc|ccc}
    \toprule
    \quad & \multicolumn{3}{c|}{\textbf{ESTER}} & \multicolumn{3}{c}{\textbf{TORQUE}}\\
    \toprule
    \textbf{Model} & $F_{1}^{T}$ & $HIT@1$ & $EM$ & $F_{1}$ & $EM$ & $C$\\
    \midrule
    TranCLR 
    & \textbf{74.7} & 80.4 & \textbf{18.3} & / & / & / \\
    Ro-L 
    & 68.8 & 66.7 & 16.7 & 75.7 & 50.4 & 36.0\\
    Ro-L I-O & 73.7 & 77.4 & 15.3 & 75.8 & 50.8 & 36.1\\
    Ro-L PR & 74.0 & \textbf{81.7} & \textbf{18.3} & \textbf{76.2} & \textbf{50.8} & 37.5\\
    \midrule
    Ro-L PR (-trig) & 74.0 & 80.8 & 17.6 & 76.1 & 50.7 & \textbf{37.7}\\
    \bottomrule
  \end{tabular}}
    \caption{Extractive QA results on the ESTER and TORQUE datasets. TranCLR results are taken from \cite{lu2022event}. Ro-L refers to RoBERTa-large, the backbone encoder of all the reported models. Ro-L I-O identifies models to fine-tuned with the unified "I-O" tagging schema and balanced weights. Ro-L PR indicates the fine-tunined RoBERTa-large models using the proposed posterior regularization (PR) mechanism.  Ro-L PR (-trig) refers to the ablation setting that performs answer extraction without using the event triggers during inference.
    }
    \vspace{-12pt}
  \label{tab:extQA}
\end{table}

\subsection{Experimental Results}
\paragraph{Extractive QA} 
Table \ref{tab:extQA} reports the extractive QA results. We implemented a unified "I-O" tagging schema with balancing weights for both datasets, as shown in Eq. (\ref{eq:extl}). This is slightly different from the \cite{han2021ester} approach based on the  "B-I-O" tagging schema with a balancing weight for the positive labels "B" and "I", and the binary "I-O" tagging schema of \cite{ning2020torque} without any balancing weight. The "Ro-L I-O" model refers to the unified baseline based on the binary schema, in which the balancing weights are empirically set to 4.0 and 1.0 for the ESTER and TORQUE dataset, respectively. 
The consistent improvements across different metrics and datasets show the broad benefits of the proposed regularization approach. 

On the ESTER dataset, the introduction of the "\mbox{I-O}" tagging schema alone brings an absolute improvement of $4.9\%$ in $F_{1}^{T}$ and $10.7\%$ in $HIT@1$, with a slightly worse $EM$ score. 
Applying the proposed posterior regularization (PR), we see a further increase in performance across all metrics. \jr{Compared with TranCLR \cite{lu2022event}, which encodes the token-level event knowledge through contrastive learning, our sentence-level event knowledge constraints can incorporate higher-level linguistic information, and gives a higher $HIT@1$ score.} 
While on TORQUE, our proposed posterior regularization mechanism increases the token-level $F_1$ and $C$ scores by 0.4\% and 1.4\%, respectively. \jr{The improvement on F1 is marginal since the answer spans here are only short event triggers, as shown in Figure \ref{fig:sample}}. 
\gab{While all baselines assume the event trigger information is given during inference, we further conduct an ablation study "Ro-L PR (-trig)" by removing the information of event triggers, and predicting event scores over all tokens in Eq. (\ref{eq:etype}). The performance remains nearly the same compared with the model where the actual event trigger positions are known
.}

\begin{table}[t]
\resizebox{\columnwidth}{!}{%
  \begin{tabular}{lccc}
    \toprule
    \textbf{Model} & $F_{1}^{T}$ & $HIT@1$ & $EM$\\
    \midrule
    TranCLR \cite{lu2022event} 
    & \textbf{74.2} & 86.4 & 25.6 \\
    Unified-T5-Large \cite{han2021ester} 
    & 66.8 & \textbf{87.2} & 24.4 \\
    Unified-T5-Large PR ($\tau_1=0.5$, $\tau_2=1.5$) & 71.4 & 86.7 & \textbf{26.0} \\
    \bottomrule
  \end{tabular}}
    \caption{Generative QA results on the ESTER dataset. TranCLR results are taken from \cite{lu2022event}. Unified-T5-Large refers to fine-tuned T5-Large model in UnifiedQA pipeline \cite{khashabi2020unifiedqa}. Unified-T5-Large PR refers to fine-tuning T5-Large model with our proposed posterior regularization.}
    \vspace{-12pt}
  \label{tab:genQA}
\end{table}

\paragraph{Generative QA} Results of generative QA are reported in Table \ref{tab:genQA}. Our PR approach gains $4.6\%$ and $1.6\%$ absolute improvements on unigram level $F_{1}^{T}$ and $EM$ scores. \jr{We obtain higher $HIT@1$ and $EM$ scores compared with TranCLR, while a lower token-level $F_1$ score.} The values of the reward \& penalty temperatures $\tau_1$ and $\tau_2$ have some impact on the QA task. The grid-search results are reported in Appendix \ref{sec:gridsearch}.

\begin{figure}[t]
  \centering
  \includegraphics[width=\linewidth]{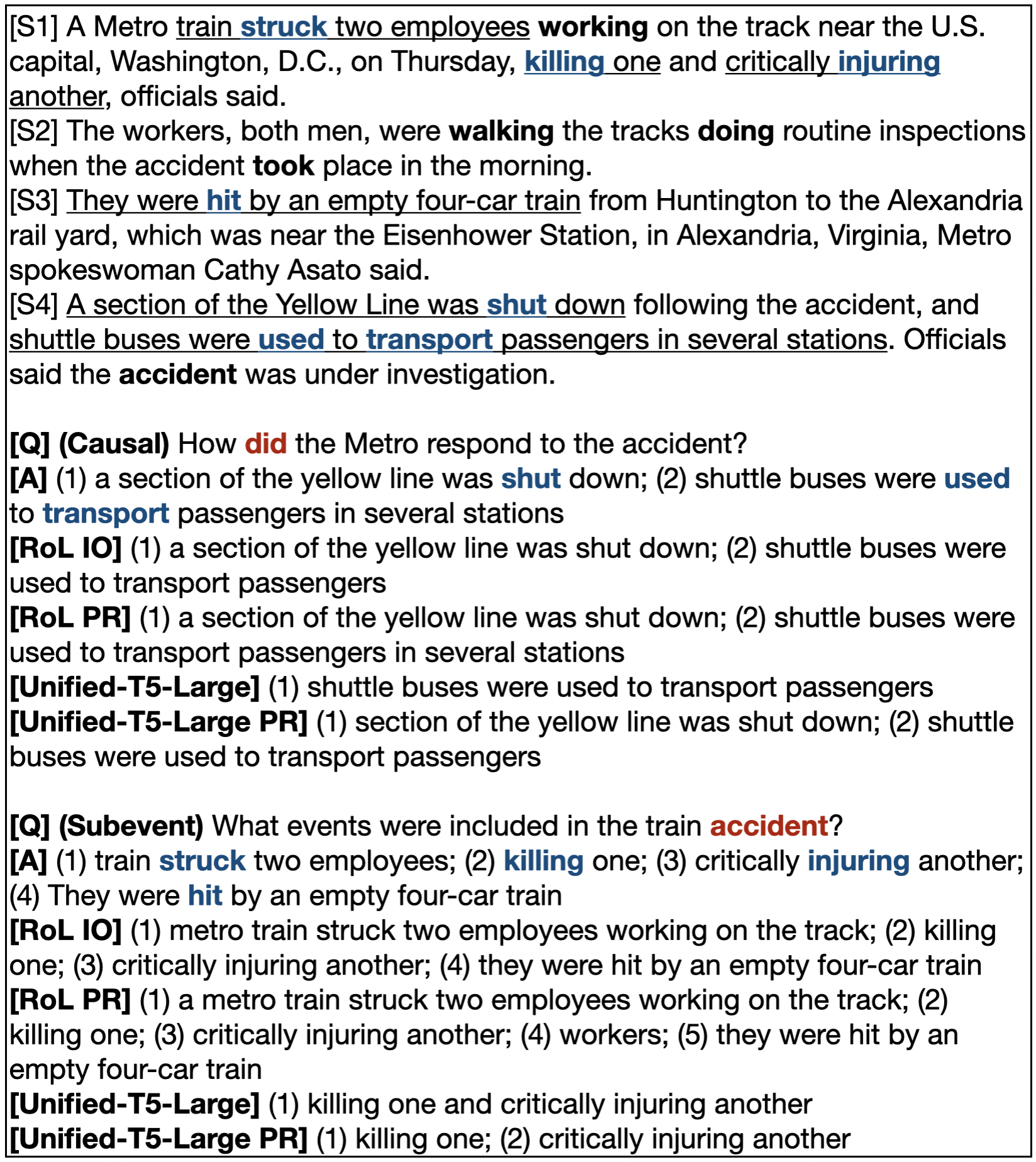}
  \caption{Answers generated by different QA models on two types of typical event-centric questions.}
  \label{fig:results}
\end{figure}

\paragraph{Case Study and Error Analysis} We select two questions for case analysis in Figure \ref{fig:results}. The first case is a "\emph{Causal}" question. Extractive models perform better than generative ones. Specifically, RoBERTa-large PR model extracts all answers correctly, while the RoBERTa-large baseline misses several tokens at the end of the second answer. In the generative QA setting, the Unified-T5-Large misses the first answer and only generates the second one, while the Unified-T5-Large PR correctly generates both answers. The second question is from the hardest "\emph{Sub-event}" group. As in the previous case, extractive models are more accurate. Our RoBERTa-large PR model mistakenly extracts a "\emph{workers}" answer, suggesting that the model incorrectly predicted a high relevance score for sentence 2. In this case, both generative models produce partial answers, although our PR model accurately splits the span into two answer events.

\section{Conclusions}
We propose a simple yet effective mechanism to incorporate event knowledge via posterior regularization for event-centric QA. We designed knowledge constraints based on the event trigger annotations and used them to regularize the answer probabilities generated by the backbone models. In extractive QA, the sentence-level event knowledge constraint was set up to regularize the answer probability depending on whether a sentence contained the answer events. In generative QA, token-level regularization terms reward the generation of target events and penalize the prediction of irrelevant events. The experiments showed the effectiveness of our regularization mechanism on various event-centric QA datasets and across different metrics.

\section*{Limitations}
\jr{Considering the different mechanisms for answer extraction and generation, we design sentence-level posterior constraints for extractive QA and token-level posterior constraints for generative QA, respectively. Although the two settings are formulated in a unified framework in Eq. (\ref{eq:G-function}), the $G$ function needs to be designed separately. 
Reinforcement learning could be applied in the future to automatically learn meta $G$ function in connection with prior distribution, knowledge constraints and posterior distribution \cite{zoph2016neural}.}

\jr{The experimental results show the effectiveness of our methodology on various event-centric questions involving different event types and question formats. Nevertheless, our models only consider the explicitly annotated event triggers and reference answer information, and thus only obtain marginal improvements on the TORQUE dataset containing only single-token answers. It is worth exploring implicit event-related information (e.g. event arguments \cite{xiang2019survey} and event relations \cite{liu2021survey}) or external event knowledge (e.g. event knowledge graph \cite{gottschalk2019eventkg}) for the event-centric QA task.}

\bibliography{acl2023}
\bibliographystyle{acl_natbib}

\appendix

\setcounter{table}{0}
\renewcommand{\thetable}{A\arabic{table}}

\section{Hyperparameters}
\label{sec:hyperparams}
We follow the original dataset papers \cite{han2021ester,ning2020torque} for most of the following hyperparameters settings.

For extractive QA, the hidden size of RoBERTa-large model is 1,024. The vocabulary size is 50,265. The attention layer used in Eq. \ref{eq:attention} is a 1-block vanilla 8-heads MHA network. The batch size, accumulation steps, and random seed are (16, 2, 23) for the ESTER and (8, 2, 24) for the TORQUE datasets, respectively. The optimizer is \textit{BertAdam} \cite{devlin2018bert} with $\beta_1\,$= 0.9, $\beta_2\,$= 0.999, and $\epsilon\,$= 1e-6. Except for the parameters in the normalization layer, all other trainable parameters are fine-tuned with 0.95 decaying rate, with the learning rate 1e-5. The task hyperparameter $\lambda_1$ in Eq. \ref{eq:extqa} is set to 0.1. The balancing weight in Eq. \ref{eq:extl} is set to 4.0 and 1.0 for the ESTER and TORQUE datasets, respectively. It takes 1.0 and 4.5 hours to fine-tune RoBERTa-large on the ESTER and TORQUE datasets, respectively, on two RTX 6000 GPUs.

For generative QA, the hidden size of UnifiedQA-T5-large is 1,024 and its vocabulary size is 32,128. The random seed is 5. We use the same aforementioned GPUs, with the batch size of 2 and the accumulation steps 3 during training. The optimizer and decaying strategy remain the same as extractive QA, with an increased learning rate 5e-5. The task hyperparameter $\lambda_2$ in Eq. \ref{eq:genqa} is set to 0.1. The best temperatures, in Eq. \ref{eq:temp}, are $\tau_1=0.5$ and $\tau_2=1.5$. It takes 3.5 hours to fine-tune our models on the ESTER dataset.

\section{$\tau_1$ and $\tau_2$ grid-search in Generative QA}
\label{sec:gridsearch}

The grid search results in Table \ref{tab:gridsearch} show that a lower or a higher reward weight of answer events, $\tau_1$, has negative impacts on the $F_{1}^{T}$ score, while a lower or a higher penalty weight $\tau_2$ of irrelevant events would decrease the $EM$ score.

\begin{table}[H]
\resizebox{\columnwidth}{!}{%
  \begin{tabular}{lccc}
    \toprule
    \textbf{Model} & $F_{1}^{T}$ & $HIT@1$ & $EM$\\
    \midrule
    Unified-T5-Large PR ($\tau_1=0.5$, $\tau_2=1.5$) & \textbf{71.4} & \textbf{86.7} & \textbf{26.0} \\
    \midrule
    Unified-T5-Large PR ($\tau_1=0.5$, $\tau_2=1.0$) & 70.6 & 86.1 & 23.0 \\
    Unified-T5-Large PR ($\tau_1=0.5$, $\tau_2=2.0$) & 70.3 & \textbf{86.7} & 24.0 \\
    Unified-T5-Large PR ($\tau_1=0.0$, $\tau_2=1.5$) & 69.4 & 85.1 & 25.0 \\
    Unified-T5-Large PR ($\tau_1=1.0$, $\tau_2=1.5$) & 69.3 & 84.7 & 23.4 \\
    \bottomrule
  \end{tabular}}
    \caption{Generative QA results on the ESTER dataset. Unified-T5-Large refers to fine-tuned T5-Large model in UnifiedQA pipeline \cite{khashabi2020unifiedqa}. Unified-T5-Large PR refers to fine-tuning with the new posterior regularization, with a breakdown of the gridsearch over ($\tau_1$, $\tau_2$) for the knowledge constraint defined in Eq. (\ref{eq:temp}).}
    \vspace{-12pt}
  \label{tab:gridsearch}
\end{table}

\end{document}